# CNN-based Two-Stage Parking Slot Detection Using Region-Specific Multi-Scale Feature Extraction

Quang Huy Bui and Jae Kyu Suhr

Department of Intelligent Mechatronics Engineering, Sejong University, Seoul 05006, Republic of Korea

E-mail: quanghuy2907@sju.ac.kr, jksuhr@sejong.ac.kr

*Abstract*

Autonomous parking systems start with the detection of available parking slots. Parking slot detection performance has been dramatically improved by deep learning techniques. Deep learning-based object detection methods can be categorized into one-stage and two-stage approaches. Although it is well-known that the two-stage approach outperforms the one-stage approach in general object detection, they have performed similarly in parking slot detection so far. We consider this is because the two-stage approach has not yet been adequately specialized for parking slot detection. Thus, this paper proposes a highly specialized two-stage parking slot detector that uses region-specific multi-scale feature extraction. In the first stage, the proposed method finds the entrance of the parking slot as a region proposal by estimating its center, length, and orientation. The second stage of this method designates specific regions that most contain the desired information and extracts features from them. That is, features for the location and orientation are separately extracted from only the specific regions that most contain the locational and orientational information. In addition, multi-resolution feature maps are utilized to increase both positioning and classification accuracies. A high-resolution feature map is used to extract detailed information (location and orientation), while another low-resolution feature map is used to extract semantic information (type and occupancy). In experiments, the proposed method was quantitatively evaluated with two large-scale public parking slot detection datasets and outperformed previous methods, including both one-stage and two-stage approaches.

## 1 Introduction

As a result of the growing interest in autonomous driving, autonomous parking systems have gained increasing attention. Such systems have proven their role by providing drivers convenience and reducing vehicle damage (Banzhaf et al., 2017; Khalid et al., 2020; Suhr & Jung, in press). In autonomous parking, the first step is to precisely detect an available parking space. Recently, a soaring number of vehicles are equipped with vision systems that enhance the drivers' awareness of their surroundings. Some clear examples are the rearview camera and around view monitor (AVM) system, which eliminates the rear blind spot and provides 360 degrees of observation around the vehicle, respectively. This tendency has led to the significant development of vision-based parking slot detection.

The initial methods for vision-based parking slot detection are based on hand-crafted features. These methods extract line or corner features from images and combine them using geometric rules to find parking slots. Although they have shown noticeable performances, the inconvenience of designing adequate geometric rules and the fragility of those rules to various environmental conditions have been revealed as their significant drawbacks. In recent years, with the rise of the deep learning, convolutional neural network (CNN) has made considerable breakthroughs in numerous object detection tasks. CNN-based general object detection methods can be categorized into two main approaches: two-stage and one-stage. The two-stage approach consists of one step to generate region proposals and the other step to classify the objects inside those regions and refine their bounding boxes. Region-based CNN (RCNN) (Girshick et al., 2015), Fast RCNN (Girshick, 2015), Faster RCNN (Ren et al., 2015), RFCN (Dai et al., 2016), and Mask-RCNN (He et al., 2017) are representative methods for this approach. On the other hand, the one-stage approach directly acquires bounding boxes for the objects along with their classes without generating region proposals. You only look once (YOLO) (Redmon et al., 2016), YOLOv2 (Redmon & Farhadi, 2017), YOLOv3 (Redmon & Farhadi, 2018), YOLOv4 (Bochkovskiy et al., 2020) single shot multibox detector (SSD) (Liu et al., 2016), and RetinaNet (Lin et al., 2017) are representative methods for this approach. Through various applications, the two-stage approach has shown a high detection performance with a slow processing speed, while the one-stage approach has shown a moderate detection performance with a fast processing speed. Witnessing the success of CNN-based object detection, many research works have been conducted to utilize it for parking slot detection tasks.

Similar to general object detection, CNN-based parking slot detection methods can be categorized into two approaches: two (or multi)-stage and one-stage. In multi-stage parking slot detection methods, the first stage generates region proposals by finding two or four corners of the parking slots (Li, Cao, Yan, et al., 2020; Zinelli et al., 2019) or by combining parts of the parking slots found by CNNs using geometrics rules (Zhang et al., 2018; Huang et al., 2019; Jang & Sunwoo, 2019; Jiang et al., 2020; Min et al., 2021). Then, the following stages refine the positions or classify types and occupancies of the parking slots by extracting features of the region proposals from the corresponding regions of the feature map or input image. On the other hand, one-stage parking slot detection methods directly acquire all information of the parking slot such as location, orientation, type, and occupancy in a single step without generating region proposals (Li, Cao, Liao, et al., 2020; Suhr & Jung, 2021). Even though the two-stage detection approach has been known to outperform the one-stage detection approach in general object detection tasks, their performances have been reported to be similar in parking slot detection tasks. The state-of-the-art one-stage parking slot detector has shown a slightly better performance than the two-stage parking slot detectors (Li, Cao, Liao, et al., 2020; Suhr & Jung, 2021). We consider this is because the two-stage approach has not yet been adequately specialized for parking slot detection tasks.

Therefore, this paper proposes a highly specialized two-stage parking slot detector that uses region-specific multi-scale feature extraction. In the first stage, the proposed method finds the entrance of the parking slot as a region proposal by predicting its location, orientation, and length. It is unlike the previous methods that adopt an upright rectangle (Li, Cao, Yan, et al., 2020) or four corners of the parking slot (Zinelli et al., 2019) as a region proposal. In the second stage, this method uses a region-specific feature extraction method that extracts features only from the specific regions of the feature map that most contain the desired information. For instance, features for predicting the location and orientation of the parking slot are separately extracted from only the specific regions that most contain the corresponding information. This is possible because the parking slot is a planar rigid object on the ground plane and captured in an AVM image after removing perspective distortion. It is unlike the previous methods that extract the features of the entire region proposal from the feature map (Zinelli et al., 2019) or crop the whole area of the region proposal from the input image (Li, Cao, Yan, et al., 2020). In addition, the proposed method utilizes multi-resolution feature maps to increase both positioning and classification accuracies. It uses a high-resolution feature map for extracting detailed information (location and orientation) and a low-resolution feature map for extracting semantic information (type and occupancy). Finally, from the extracted features, the proposed method refines the locations and orientations of the parking slots and classifies their types and occupancies. In experiments, the proposed method was quantitatively evaluated with two large-scale public parking slot detection datasets and outperformed previous methods, including both one-stage and two-stage approaches. The contributions of this paper can be summarized as follows:

1) It suggests an effective way to apply the two-stage general object detection to the parking slot detection task.
2) It proposes a region-specific multi-scale feature extraction that increases both detection performance and positioning accuracy.
3) It presents quantitative evaluation results using two large-scale public datasets and shows that the proposed method gives a state-of-the-art performance.

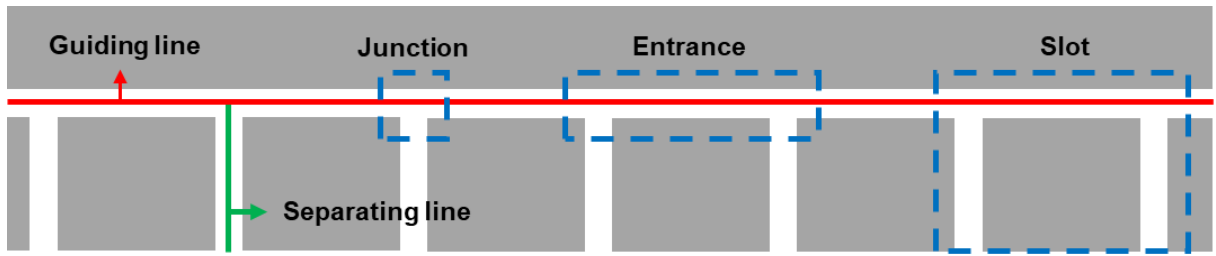

Fig. 1. Terminologies for parking slot markings.

## 2  Related works

Previous vision-based parking slot detection methods can be categorized into hand-crafted feature-based and deep learning-based (or CNN-based). Since these methods exploit parking slot markings on the ground, terminologies for the parking slot markings are briefly introduced in Fig. 1. In this figure, the guiding line segregates the parking slots from the roadway, and separating lines divide individual parking slots. Junctions are the intersections of the guiding line and separating lines, and the entrance of a parking slot is the segment between two adjacent junctions. A parking slot is formed by the entrance and a pair of separating lines connecting to it.

Hand-crafted feature-based methods detect parking slots by extracting manually designed features of the parking slot and combining them using traditional rule-based techniques. Since this paper concentrates mainly on the deep learning-based methods, the hand-crafted feature-based methods are briefly introduced. Based on the type of extracted features, the hand-crafted feature-based methods can be categorized into line-based and junction-based. The line-based methods first find the guiding lines and separating lines and then group them to generate parking slots. Various techniques have been employed to detect and combine line features. For line detection, Hough transform (Jung et al., 2006; Hamada et al., 2015), Radon transform (Wang et al., 2014; Kim et al., 2020), or random sample consensus (RANSAC) algorithm (Du & Tan, 2014; Lee & Seo, 2016; Zong & Chen, 2018; Suhr & Jung, 2018) have been utilized. For line combination, K-means clustering (Wang et al., 2014), grouping based on predetermined distances and parallel and perpendicular properties (Hamada et al., 2015; Chen & Hsu, 2017; Suhr & Jung, 2016, 2018) have been used. Different from the line-based methods, the junction-based methods first find junctions of the parking slots and then pair them to generate parking slot candidates. For junction detection, Harris corner detector (Jung et al., 2009; Suhr & Jung, 2012, 2013) and Viola-Jones detector (Li et al., 2017) have been applied. The detected junctions are paired by various geometric rules based on their types, locations, and orientations. Once parking slots are detected by the line-based or junction-based methods, their occupancies are then classified. To this end, difference-of-Gaussians-based histogram with linear discriminant analysis (LDA) classifier (Houben et al., 2013), Canny edges with naïve Bayes classifier (Chen & Hsu, 2017), color histogram with support vector machine (SVM) classifier (Kim et al., 2020), ultrasonic sensor-based occupancy grid (Suhr & Jung, 2013, 2016) have been exploited.

As CNN-based object detection has shown significant results in recent years, various research works have been done to apply this technique to the parking slot detection task. CNN-based parking slot detection methods can be categorized into two approaches: multi-stage and one-stage. The first multi-stage parking slot detection method applying deep learning technique was proposed by Zhang et al. (2018). The first stage of this method finds junctions using YOLOv2 and its second stage generates parking slot candidates by combining the junctions using geometric rules. Finally, a CNN-based classifier verifies the candidates whose orientations are determined by a template matching technique in the last stage. Similarly, Huang et al. (2019) customized a CNN to find locations, orientations, and types of junctions and then grouped them using geometric rules to generate parking slot candidates. This method can handle perpendicular and parallel parking slots. The method proposed by Li, Cao, Yan, et al. (2020) detects junctions and entrances using YOLOv3 with upright bounding boxes and finds parking slots by means of geometric rules and relation between the detected junctions and entrances in the first stage. Its second stage separately crops the regions of the parking slots from the input image and forwards them to an additional CNN for occupancy classification. This method cannot perform well when the ego-vehicle is inside a parking slot and can inaccurately estimate orientations due to the predefined orientations for slanted parking slots. Jang and Sunwoo (2019) and Jiang et al. (2020) proposed methods that extract the marking lines and junctions of parking slots using semantic segmentation techniques in the first stage. They generate parking slots using extracted lines and junctions along with geometric rules and classify their occupancies based on the semantic segmentation results in the second stage. All aforementioned methods have shown the potential of deep learning techniques in parking slot detection tasks. However, they cannot be trained end-to-end due to the manual selection of geometric rules and associated parameters, which is inconvenient and complicated to set. To overcome this limitation and benefit the training process, end-to-end trainable methods have been proposed. Zinelli et al. (2019) presented the first end-to-end trainable parking slot detection method utilizing anchor-free faster R-CNN (Zhong et al., 2019). The first stage of this method roughly estimates four corners of the parking slot as a region proposal. RoIAlign (He et al., 2017) is then used to extract features from the proposed region for location refinement and occupancy classification in the second stage. Trying to apply a general object detection to the parking slot detection task, this method, however, shows clear limitations of detection performance and positioning accuracy because it uses the general object detector without sufficient modification. Another end-to-end trainable two-stage parking slot detection method was proposed by Do and Choi (2020). In the first stage of this method, the context recognizer predicts the common type and orientation of all parking slots in the input image. Then, the parking slot detector estimates the exact positions of the parking slots using rotated anchor boxes in the second stage. Although this method can obtain all information of the parking slots, including location, orientation, type, and occupancy, it requires a high computational cost due to the use of two separate backbone networks and handles only the cases where all parking slots in the input images have the same type and orientation. Min et al. (2021) proposed a three-stage parking slot detection method. It finds junctions and extracts their features in the first stage and aggregates the junctions to generate parking slot candidates using an attentional graph neural network in the second stage. Finally, those candidates are verified based on a multilayer perceptron in the last stage. This method is limited in dealing with slanted parking slots due to the absence of orientation information extraction.

Since multi-stage parking slot detection methods, in general, are mediocre in terms of inference speed, one-stage parking slot detection methods have also been suggested. Li, Cao, Liao, et al. (2020) introduced a one-stage parking slot detection method focusing on locating the entrance of the parking slot. This method predicts the location, orientation, and type of the parking slot entrance using a customized CNN. Although it shows a fast inference speed with an adequate detection performance, it provides no occupancy information and unsatisfactory orientation accuracy due to the predefined orientations for slanted parking slots. Suhr and Jung (2021) suggested another one-stage parking slot detection method. This method simultaneously extracts global information (rough location, type, and occupancy of the parking slot) and local information (precise location and orientation of junctions) and combines them to provide final parking slots. This method achieves a high detection performance requiring only a low computational cost while providing all information of the parking slot (location, orientation, type, and occupancy).

As a thorough literature review, it is observed that currently, for parking slot detection tasks, one-stage detection methods slightly outperform multi-stage detection methods in both aspects: detection performance and positioning accuracy. This is unlike general object detection tasks, where the two-stage approach outperforms the one-stage approach. We consider one of the main reasons for this is that the two-stage approach has not yet been adequately specialized for parking slot detection tasks. Therefore, this paper proposes a highly specialized two-stage parking slot detector. In experiments, it has been revealed that the adequately designed two-stage parking slot detection method outperforms the one-stage parking slot detection methods.

## 3 Proposed method

### 3.1 Overall architecture

This paper proposes a novel two-stage parking slot detection method using region-specific multi-scale feature extraction. The proposed method roughly locates parking slot entrances using the region proposal network (RPN) in the first stage and precisely estimates positions and properties of parking slots using the slot detection network (SDN) and slot classification network (SCN) in the second stage. Fig. 2 illustrates the overall architecture of the proposed method. An input AVM image, as in Fig. 2(a), is inserted into the backbone network for feature maps extraction. This paper tried several backbone networks and selected DenseNet121 (Huang et al., 2017), whose performance has been proven in various applications. After acquiring the feature maps, the RPN with one convolutional layer is applied to the low-resolution feature map to generate approximate positions of parking slot entrances as region proposals. Fig. 2(b) shows the output of the RPN, where solid black lines and arrows indicate the entrances and orientations of the parking slots, respectively. Once region proposals are generated, this paper applies the region-specific multi-scale feature extraction to estimate the positions and properties of parking slots more accurately. Rather than utilizing features of the entire parking slot, the region-specific approach extracts features from only the regions that most contain the desired information. Magenta and yellow squares in Fig. 2(c) are the specific regions used to extract features for estimating the locations and orientations of the parking slot, respectively. These regions include junctions and separating lines, and thus contain rich locational and orientational information. Purple squares in Fig. 2(d) are the specific regions used to extract features for type and occupancy classification. These regions include the center areas of parking slots that contain overall shape and texture information. In addition, multi-resolution feature maps are utilized to enhance positioning and classification performances. The high-resolution feature map, containing more detailed information, is used to extract features to estimate the locations and orientations of parking slots, while the low-resolution feature map, containing more semantic information, is used to extract features to classify their types and occupancies. After obtaining the features using the proposed region-specific multi-scale feature extraction, the SDN with a set of fully connected layers is applied to estimate precise positions of the parking slots, as marked with black lines in Fig. 2(e). Concurrently, the SCN with a set of fully connected layers is applied to estimate types and occupancies of the parking slots. Fig. 2(f) shows the output of the SCN where blue solid, red solid, and green dashed rectangles indicate vacant slanted, vacant parallel, and occupied perpendicular parking slots, respectively. The proposed method determines the final parking slots by combining their positions, types, and occupancies, as illustrated in Fig. 2(g).

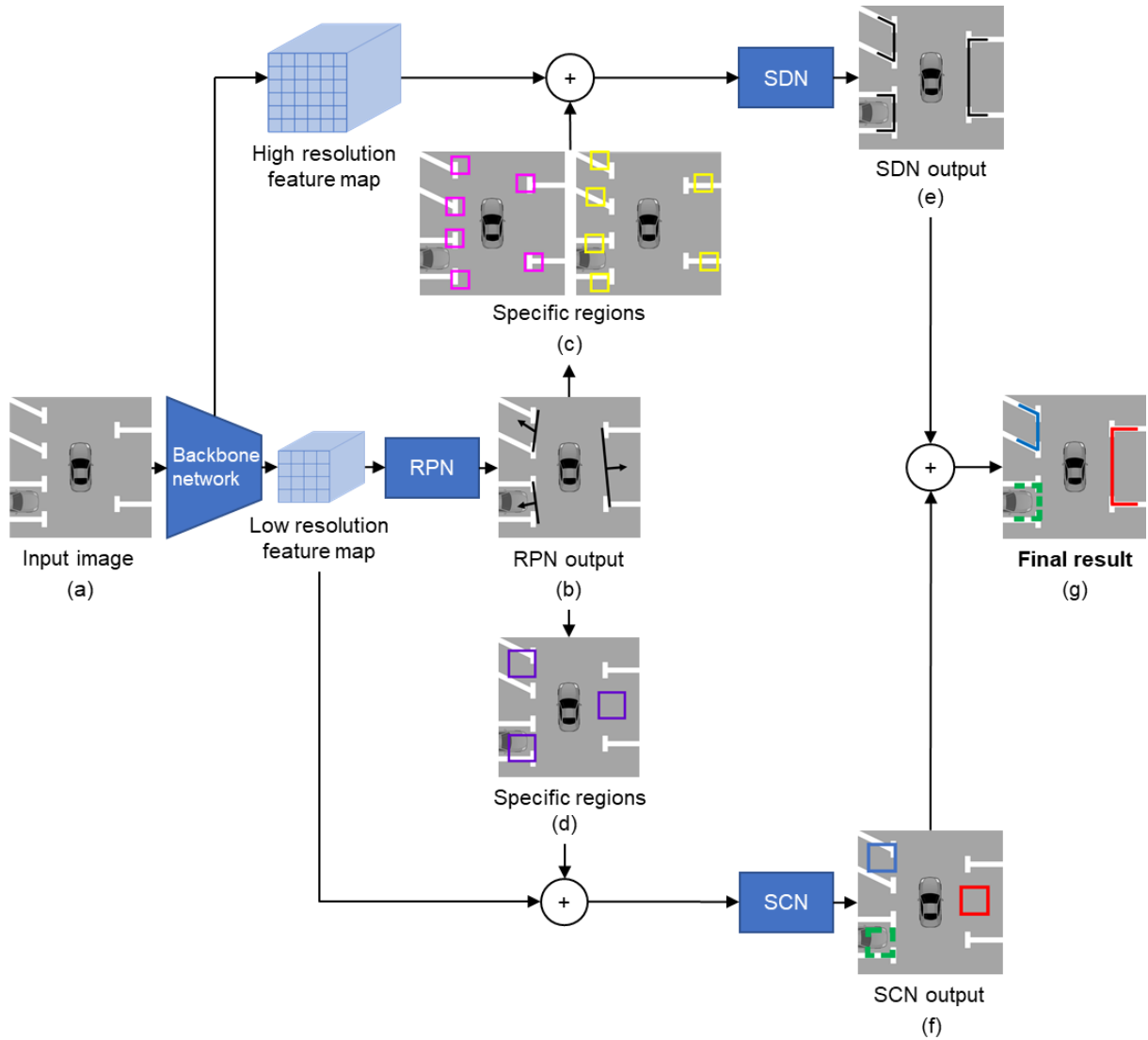

Fig. 2. Overall architecture of the two-stage method utilizing region-specific multi-scale feature extraction.

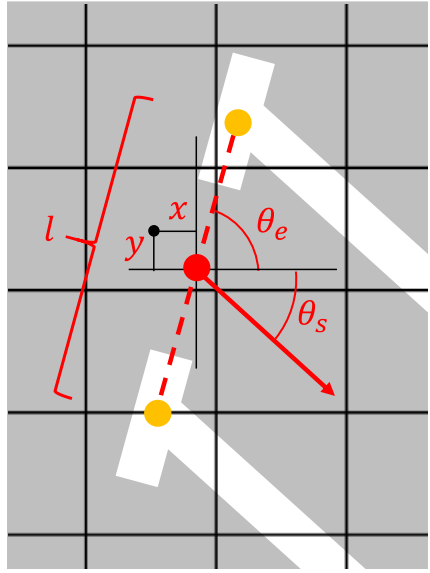

Fig. 3. Representation of the parking slot entrance using its center location $(x, y)$, orientation $(\cos \theta_e, \sin \theta_e)$, length $(l)$, and parking slot orientation $(\cos \theta_s, \sin \theta_s)$.

### 3.2 Region proposal network

The proposed method generates the parking slot entrance as a region proposal, unlike previous methods that capture the whole parking slot using a parallelogram (Li, Cao, Yan, et al., 2020), quadrilateral (Zinelli et al., 2019), or rotated rectangle (Do & Choi, 2020). This is because AVM images do not usually include the whole parking slot, and the parking slot entrance itself contains enough information for cars to start parking. This procedure also avoids geometric rule-based approaches (Zhang et al., 2018; Huang et al., 2019; Jang & Sunwoo, 2019; Jiang et al., 2020) that hinder end-to-end training. To represent the parking slot entrance, we considered two approaches suggested by Li, Cao, Liao, et al. (2020) and Suhr and Jung (2021). The former uses the location and orientation of the entrance center, and the latter uses the locations of the paired junctions. Based on the experimental comparison, this paper modifies the approach suggested by Li, Cao, Liao, et al. (2020) and represents the parking slot entrance by its center location $(x, y)$, orientation $(\cos\theta_e, \sin\theta_e)$, length $(l)$, and the orientation of the parking slot $(\cos\theta_s, \sin\theta_s)$ as shown in Fig. 3.

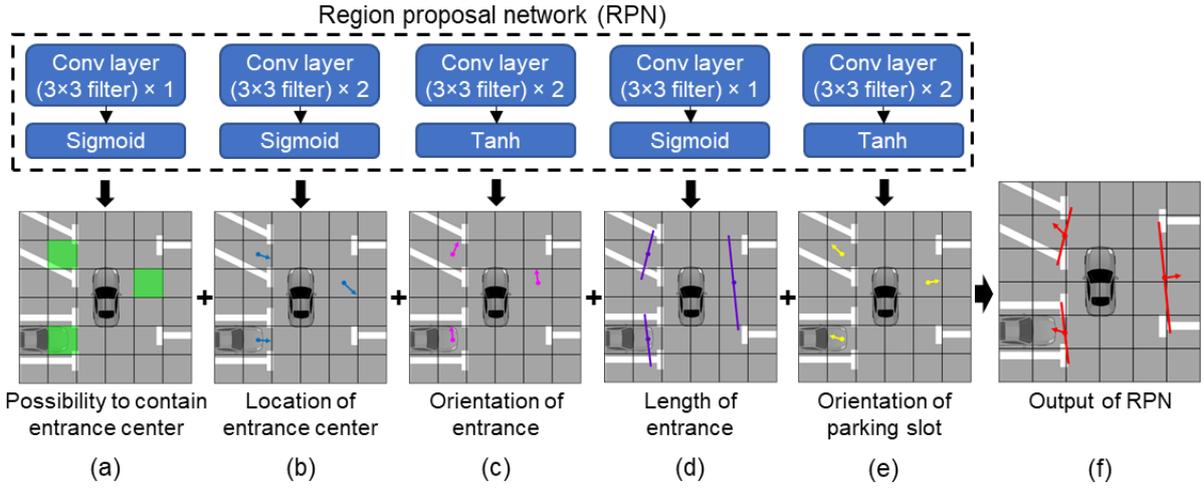

Fig. 4. Region proposal network (RPN) and detailed information obtained from it.

Fig. 4 gives a detailed description of the RPN. In the RPN, one convolutional layer with eight $3 \times 3$ filters is applied to the low-resolution feature map produced by the backbone network, as illustrated at the top of Fig. 4. The spatial dimension of the RPN's output is $h \times w$. This means that the input image is divided into a grid of $h \times w$ cells. Since one cell is responsible for at most one parking slot entrance, the cell size should be set smaller than the minimum size of the parking slots. In Fig. 4, the illustrations are intentionally depicted with a grid of $6 \times 6$ cells for ease of understanding. At the top of Fig. 4(a), the possibility that a cell contains any entrance center is estimated using one $3 \times 3$ filter followed by the sigmoid function. At the bottom of Fig. 4(a), green cells indicate the cells with high possibilities to contain entrance centers of parking slots. At the top of Fig. 4(b), the relative position from a cell center to an entrance center is calculated using two $3 \times 3$ filters followed by the sigmoid function. At the bottom of Fig. 4(b), blue arrows indicate 2D vectors connecting the cell centers to the entrance centers contained in corresponding cells. In this figure, only the results obtained from the cells containing the entrance centers are drawn. At the top of Fig. 4(c), orientations of the entrances are obtained using two $3 \times 3$ filters followed by the tanh function. Because the unit vector representing the orientation consists of values in the range of -1.0 to 1.0, the tanh function is used. At the bottom of Fig. 4(c), magenta arrows indicate 2D vectors that represent the orientations of the entrances whose centers are contained in corresponding cells. At the top of Fig. 4(d), lengths of the entrances are estimated using one $3 \times 3$ filter followed by the sigmoid function. At the bottom of Fig. 4(d), purple lines indicate the estimated lengths of the entrances. At the top of Fig. 4(e), orientations of the parking slot are calculated using two $3 \times 3$ filters followed by the tanh function. At the bottom of Fig. 4(e), yellow arrows indicate 2D vectors that represent the orientations of the parking slots whose entrance centers are contained in corresponding cells. Fig. 4(f) illustrates the output of the RPN obtained by combining all the information shown in Fig.

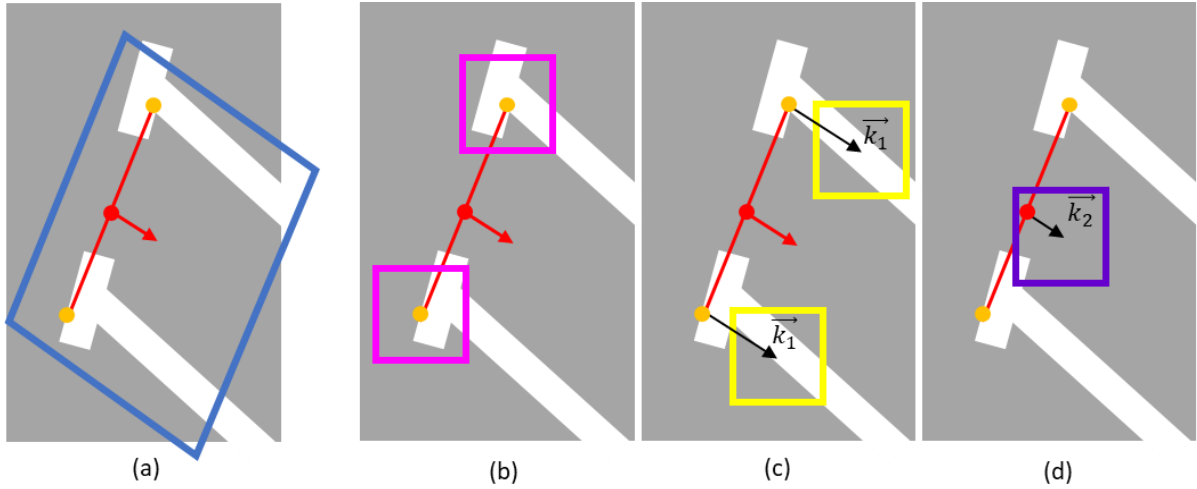

Fig. 5. (a) Parallelogram-based ROI designation; (b)-(d) Region-specific ROI designation, (b) shows location regions, (c) shows orientation regions, (d) shows type and occupancy region.

4(a)-(e). Solid red lines and arrows indicate the generated parking slot entrances and the orientations of the parking slots, respectively. Because the RPN can find multiple entrances for a single parking slot, non-maximum suppression (NMS) is utilized to remove duplicate detections based on the fact that two parking slots cannot overlap. Two entrances are considered as duplicates if their centers are closely located.

### 3.3 Region-specific multi-scale feature extraction

After generating the parking slot entrance as a region proposal, the proposed method extracts features from the region of interest (ROI) specified by the generated region proposal. General object detection methods use upright rectangles as ROIs for feature extraction (Jiao et al., 2019; Liu et al., 2020). Still, upright rectangles are inappropriate for parking slot detection because parking slots can appear with arbitrary orientations in AVM images. To tackle this problem, previous parking slot detection methods suggested other ways to designate ROIs for feature extraction, such as using parallelograms (Li, Cao, Yan, et al., 2020) or quadrilaterals (Zinelli et al., 2019). Fig. 5(a) shows a parallelogram-based ROI designation. In this figure, a blue parallelogram, inferred from the parking slot entrance, indicates the ROI for feature extraction. Since this ROI contains the whole parking slot, the features extracted from this region can predict all the information, including location, orientation, type, and occupancy. However, this approach is not optimal to designate the ROI for feature extraction in parking slot detection because specific regions of the parking slot contain features for specific information. For instance, features including locational and orientational information are mostly found in regions around junctions and separating lines, respectively. Because of this characteristic, if features are extracted from the whole parking slot region, the network can experience difficulty finding where to focus on. Our experiment has revealed that the approach using the whole region degrades the detection performance.

Therefore, to overcome the limitation of using features of the whole parking slot and enhance the detection performance, this paper proposes a region-specific ROI designation using multi-scale feature maps, called region-specific multi-scale feature extraction. The region-specific ROI designation is illustrated in Figs. 5(b)-(d). The proposed method defines only the specific regions that most contain the desired information as ROIs for feature extraction. This is possible because the parking slot is a planar rigid object on the ground plane and captured in an AVM image after removing perspective distortion, so its components, such as junctions and separating lines, can roughly be localized based on the parking slot entrance generated by the RPN. Magenta squares in Fig. 5(b) are the designated ROIs to extract features for precise location prediction. Regions around two junctions are chosen as ROIs because they contain most of the locational information. In this figure, a red line and arrow indicate the parking slot entrance generated by the RPN, and both ends of the red line are rough locations of two junctions. Yellow squares in Fig. 5(c) are the designated ROIs to extract features for precise orientation prediction. Regions around two separating lines are chosen as ROIs because they contain most of the orientational information. A purple square in Fig. 5(d) is the designated ROI to extract features for type and occupancy classification. The central region of the parking slot is used for this ROI because it contains information about the overall properties of the parking slot. The location of the ROIs in Figs. 5(c) and (d) are determined by two vectors, $\vec{k_1}$ and $\vec{k_2}$, whose directions are set to the orientation of the parking slot (red arrow), and lengths are empirically set to 50 and 32 pixels, respectively. ROIs generated by the proposed method are all upright squares. Rotated rectangles have been tried, but they did not improve the performance while increasing computational cost. Furthermore, to reduce the volume and computation of the network, this method does not crop the regions from the input image but the regions from the feature maps. This means that both its first and second stages share the backbone network, unlike some of the previous methods that crop the regions from the input image and use additional backbone networks to extract features for the second stage (Zhang et al., 2018; Li, Cao, Yan, et al., 2020).

In addition to the region-specific ROI designation, this paper suggests extracting features in different scales according to types of information. The proposed method extracts features for predicting the location and orientation from the high-resolution feature map that keeps more detailed information. On the other hand, features for predicting the type and occupancy are extracted from the low-resolution feature map that contains more semantic information. Experimental results have shown that the use of the region-specific multi-scale feature extraction remarkably increases the detection performance as well as the positioning accuracy.

Fig. 6 shows the complete operation of the proposed region-specific multi-scale feature extraction. Based on the output of the RPN shown in Fig. 6(a) with red lines and arrows indicate the entrances and orientations of the parking slots, respectively. ROIs to extract features for location, orientation, and properties (type and occupancy) are designated as shown in Fig. 6(b) with magenta, yellow, and purple squares, respectively. Once the ROIs are designated, features for location and orientation are extracted from the high-resolution feature map containing detailed information, and features for type and occupancy are extracted from the low-resolution feature map containing semantic information as shown in Fig. 6(c). When extracting features, the proposed method finds cells containing the centers of the ROIs and extracts features around these cells. 5×5 neighborhoods are used for the location and orientation, and 3×3 neighborhoods are used for the type and occupancy. A more sophisticated technique like RoIAlign (He et al., 2017) has been tried, but it did not improve the performance while increasing computational cost. The feature maps are padded with zeros during the feature extraction if a part of the ROI lays outside the feature maps.

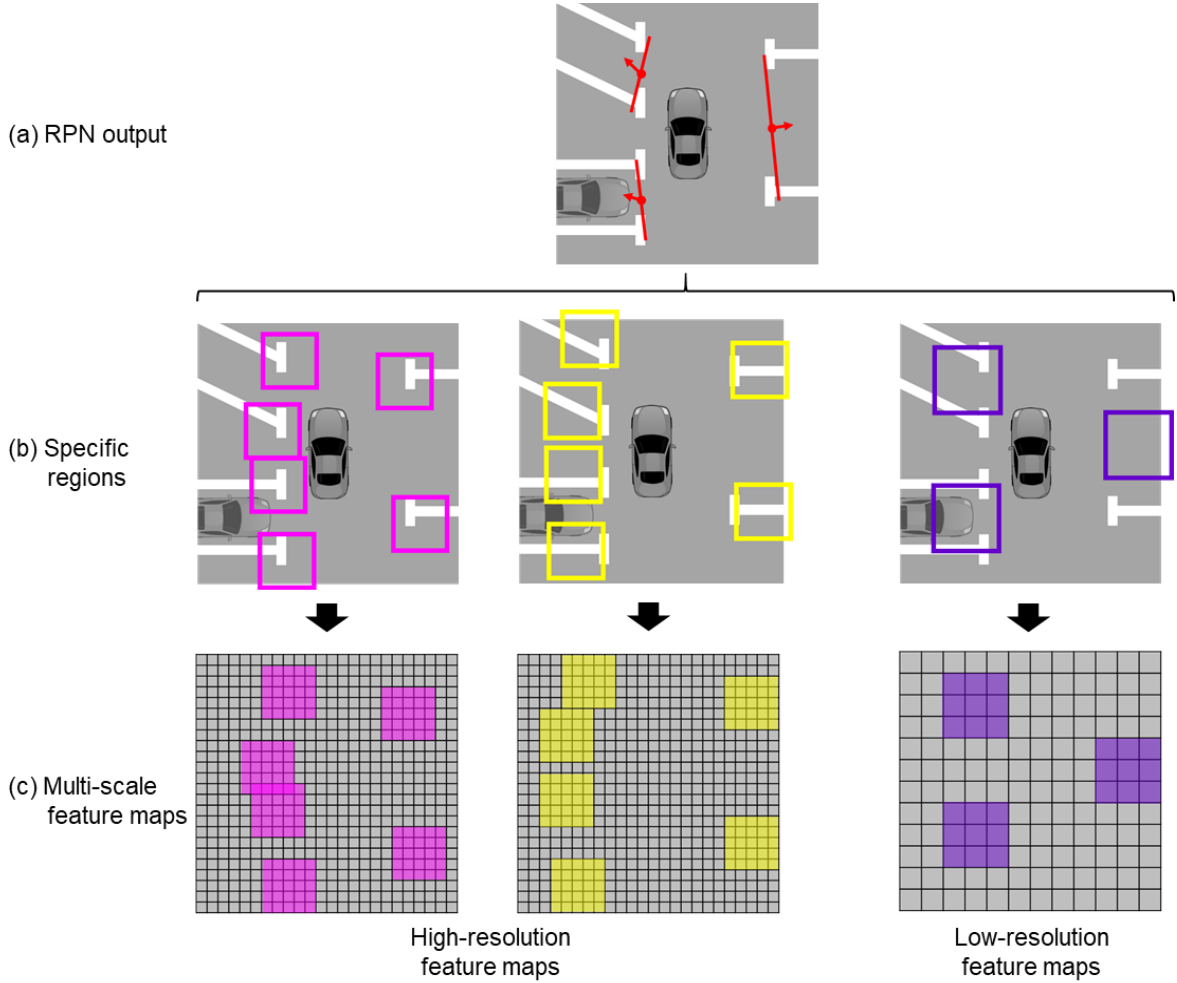

Fig. 6. Operation of the region-specific multi-scale feature extraction.

### 3.4 Parking slot detection and classification networks

Utilizing the features obtained by the proposed region-specific multi-scale feature extraction, the SDN detects the precise locations and orientations of the parking slots while the SCN classifies their types and occupancies. The top and bottom parts of Fig. 7 give detailed descriptions of the SDN and SCN, respectively. As illustrated in Fig. 7(a), for every parking slot, the region-specific multi-scale feature extractor extracts four $5 \times 5 \times 512$ tensors from the high-resolution feature map, in which two tensors are from the two junctions (in magenta squares), and the other two are from the two separating lines (in yellow squares). The SDN uses those tensors as inputs after flattening them. Fig. 7(b) shows the architecture of the SDN. Using the tensor from one magenta square, the SDN predicts three values: one for the possibility that this region contains a junction and two for the relative location from the region center to the junction. For this, two sets of fully connected layers followed by the sigmoid function are utilized. This process is separately applied to the tensors from the two magenta squares. The SDN also predicts a unit vector that describes the orientation of the separating line using the tensor from a yellow square. For this, one set of fully connected layers followed by the tanh function is utilized. This process is separately applied to the tensors from the two yellow squares. Fig. 7(c) gives a visual representation for the output of the SDN, where the red dots and arrows indicate the precisely predicted locations of the junctions and orientations of the separating lines, respectively. Similarly, as shown in Fig. 7(d), the SCN uses one $3 \times 3 \times 512$ tensor extracted from the purple square of the low-resolution feature map as an input after flattening it. From this tensor, the SCN predicts four values: one for occupancy (vacant or occupied) and three for the parking slot type (perpendicular, parallel, or slanted). For this, one set of fully connected layers followed by the sigmoid function and another set of fully connected layers followed by the softmax function are utilized, as presented in Fig. 7(e). Fig. 7(f) gives a visual representation for the output of the SCN, where the blue color and solid line indicate slanted and vacant properties, respectively. The final parking slot detection result is obtained by combining the outputs of the SDN and SCD as shown in Fig. 7(g).

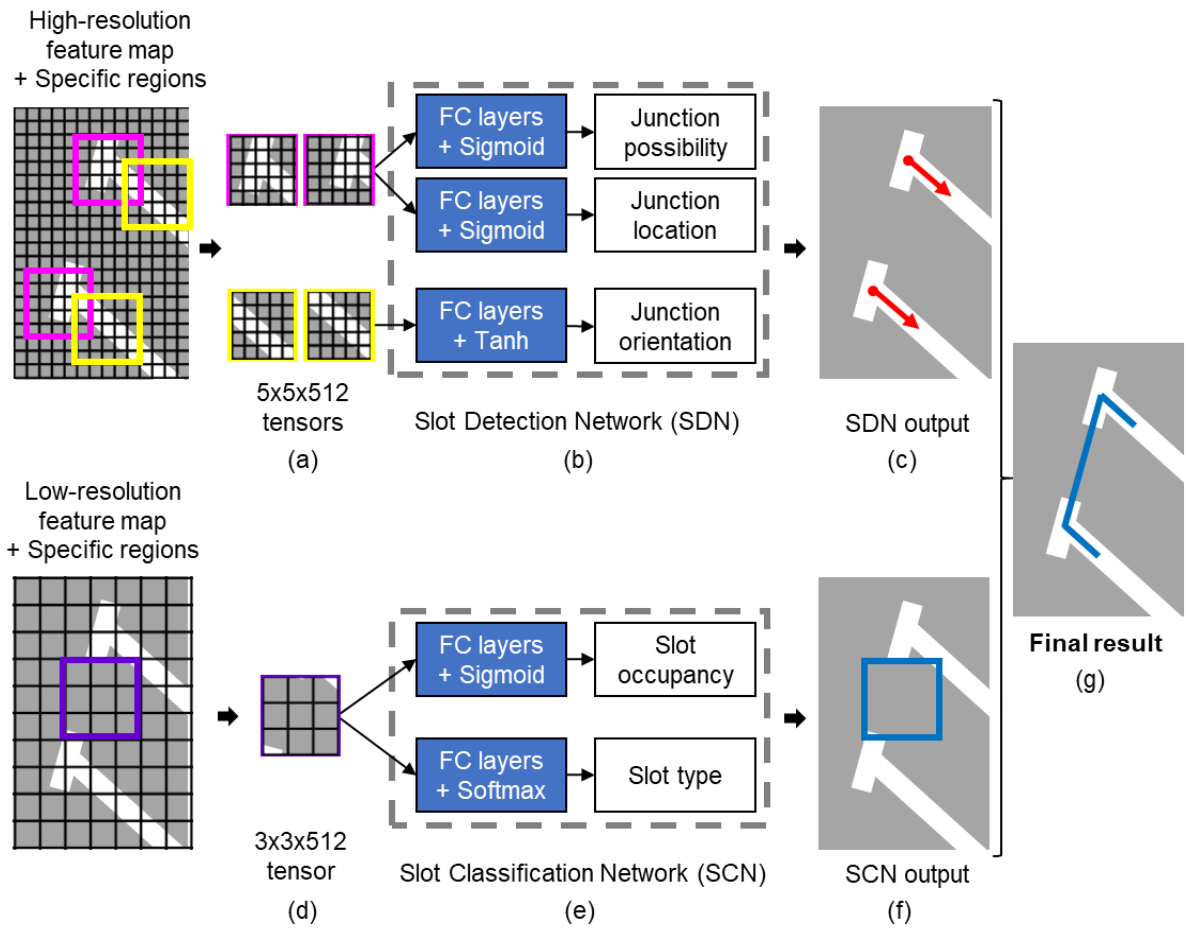

Fig. 7. Slot detection network (SDN) and slot classification network (SCN), and the detailed information obtained from them.

### 3.5 Losses

*3.5.1 Loss for the first stage*

The loss for the first stage (RPN), $loss_{first}$ is a weighted sum of five losses corresponding to five information that represents the parking slot entrance as

$$loss_{first} = w_{ep} \cdot loss_{ep} + w_{exy} \cdot loss_{exy} + w_{el} \cdot loss_{el} + w_{eo} \cdot loss_{eo} + w_{so} \cdot loss_{so} \quad (1)$$

where $w_{ep}, w_{exy}, w_{el}, w_{eo}$, and $w_{so}$ are the weights for the five losses and experimentally set. Each loss will be described in detail one by one.

The loss for the possibility that a grid cell contains an entrance center, $loss_{ep}$ is calculated as

$$loss_{ep} = \sum_{i=1}^{h \times w} \left[ I_e^i (ep_{pred}^i - ep_{true}^i)^2 + \lambda_e (1 - I_e^i)(ep_{pred}^i - ep_{true}^i)^2 \right] \quad (2)$$

where $ep_{true}^i$ is the ground truth for the possibility that the $i$-th cell includes any parking slot entrance center. This value is 1 if it includes or 0 if it does not. The input image is assumed to be divided into a grid of $h \times w$ cells. $ep_{pred}^i$ is the prediction of the network for $ep_{true}^i$. $I_e^i$ indicates whether the $i$-th cell includes any entrance center and is set to 1 if it includes or 0 if it does not. Because the number of cells that contain the entrance center is much smaller than the number of cells that do not, $\lambda_e$ is multiplied to compensate for this imbalance. It is set based on the ratio of those numbers in the training dataset.

The loss for the relative location from the cell center to the entrance center included in that cell, $loss_{exy}$ is calculated as

$$loss_{exy} = \sum_{i=1}^{h \times w} I_e^i \left[ \left\{ (ex_{pred}^i - 0.5) - \frac{ex_{true}^i}{W_{cell}} \right\}^2 + \left\{ (ey_{pred}^i - 0.5) - \frac{ey_{true}^i}{H_{cell}} \right\}^2 \right] \quad (3)$$

where $(ex_{true}^i, ey_{true}^i)$ is the ground truth for the relative location from the center of the $i$-th cell to the entrance center included in it. These values are divided by $W_{cell}$ and $H_{cell}$ to be normalized to the range of $[-0.5, 0.5]$. $W_{cell}$ and $H_{cell}$ are the width and height of the region corresponding to a single cell of the low-resolution feature map in the original image, respectively. They are 32 pixels because the backbone network includes four $2 \times 2$ pooling layers whose strides are 2. $(ex_{pred}^i, ey_{pred}^i)$ is the prediction of the network for $(ex_{true}^i, ey_{true}^i)$. Because of the sigmoid function, the predicted values are in the range of $[0, 1]$, so we subtract 0.5 from them to match their ranges with the ground truth values.

The loss for the entrance length, $loss_{el}$ is calculated as

$$loss_{el} = \sum_{i=1}^{h \times w} I_e^i \left[ el_{pred}^i - \frac{el_{true}^i}{L_{max}} \right]^2 \quad (4)$$

where $el_{true}^i$ is the ground truth for the entrance length. It is divided by $L_{max}$ to be normalized to the range of $[0, 1]$. $L_{max}$ is the maximum length of the parking slot entrance and is set based on the training dataset. $el_{pred}^i$ is the prediction of the network for $el_{true}^i$.

The loss for the orientation of the parking slot entrance, $loss_{eo}$ is calculated as

$$loss_{eo} = \sum_{i=1}^{h \times w} I_e^i \left[ (eox_{pred}^i - eox_{true}^i)^2 + (eoy_{pred}^i - eoy_{true}^i)^2 \right] \quad (5)$$

where $(eox_{true}^i, eoy_{true}^i)$ is a unit vector representing the ground truth for the orientation of the entrance whose center is included in the $i$-th cell. $(eox_{pred}^i, eoy_{pred}^i)$ is the prediction of the network for $(eox_{true}^i, eoy_{true}^i)$. These values are in the range of $[-1, 1]$ because of the tanh activation function.

The loss for the orientation of the parking slot, $loss_{so}$ is calculated as

$$loss_{so} = \sum_{i=1}^{h \times w} I_e^i \left[ (sox_{pred}^i - sox_{true}^i)^2 + (soy_{pred}^i - soy_{true}^i)^2 \right] \quad (6)$$

where $(sox_{true}^i, soy_{true}^i)$ is a unit vector representing the ground truth for the orientation of the parking slot whose entrance center is included in the $i$-th cell. $(sox_{pred}^i, soy_{pred}^i)$ is the prediction of the network for $(sox_{true}^i, soy_{true}^i)$. These values are in the range of $[-1, 1]$ because of the tanh activation function.

*3.5.2  Loss for the second stage*

The loss for the second stage, $loss_{second}$ is the sum of the loss for the SDN ($loss_{SDN}$) and the loss for the SCN ($loss_{SCN}$) as

$$loss_{second} = loss_{SDN} + loss_{SCN} \quad (7)$$

The loss for the SDN is a weighted sum of three losses corresponding to three information that represents the junction of the parking slot as

$$loss_{SDN} = w_{jp} \cdot loss_{jp} + w_{jxy} \cdot loss_{jxy} + w_{jo} \cdot loss_{jo} \quad (8)$$

where $w_{jp}, w_{jxy}$, and $w_{jo}$ are the weights for the three losses and experimentally set.

The loss for the possibility that the magenta ROIs in Fig. 7(a) include junctions, $loss_{jp}$ is calculated as

$$loss_{jp} = \sum_{i=1}^{R} [jp_{pred}^i - jp_{true}^i]^2 \quad (9)$$

where $jp_{true}^i$ is the ground truth for the possibility that the $i$-th ROI contains a junction. $R$ is the number of ROIs contained in an input image. $jp_{pred}^i$ is the prediction of the network for $jp_{true}^i$. This value is in the range of $[0, 1]$ because of the sigmoid activation function.

The loss for the relative location from the center of the magenta ROI in Fig. 7(a) to the junction included in that ROI, $loss_{jxy}$ is calculated as

$$loss_{jxy} = \sum_{i=1}^{R} I_j^i \left[ \left\{ jx_{pred}^i - \frac{jx_{true}^i}{W_{ROI}} \right\}^2 + \left\{ jy_{pred}^i - \frac{jy_{true}^i}{H_{ROI}} \right\}^2 \right] \quad (10)$$

where $(jx_{true}^i, jy_{true}^i)$ is the ground truth for the relative location from the center of the $i$-th ROI to the junction included in it. These values are divided by $W_{ROI}$ and $H_{ROI}$ to be normalized to the range of $[-0.5, 0.5]$. $W_{ROI}$ and $H_{ROI}$ are the width and height of the region corresponding to a $5 \times 5$ area of the high-resolution feature map in the original image. They are 80 pixels because the high-resolution feature map is taken from the third pooling layer of the backbone network. $I_j^i$ indicates whether the $i$-th ROI contains any junction and is set to 1 if it contains or 0 if it does not. $(jx_{pred}^i, jy_{pred}^i)$ is the prediction of the network for $(jx_{true}^i, jy_{true}^i)$. Because of the sigmoid function, the predicted value are in the range of $[0, 1]$, so we subtract 0.5 from them to match their ranges with the ground truth values.

The loss for the orientation of the separating lines in the yellow ROIs of Fig. 7(a), $loss_{jo}$ is calculated as:

$$loss_{jo} = \sum_{i=1}^{R} I_j^i \left[ \left( jox_{pred}^i - jox_{true}^i \right)^2 + \left( joy_{pred}^i - joy_{true}^i \right)^2 \right] \quad (11)$$

where $(jox_{true}^i, joy_{true}^i)$ is a unit vector representing the ground truth for the orientation of the separating line included in the $i$-th ROI. $(jox_{pred}^i, joy_{pred}^i)$ is the prediction of the network for $(jox_{true}^i, joy_{true}^i)$. These values are in the range of $[-1, 1]$ because of the tanh activation function.

The loss for the SCN is a weighted sum of two losses corresponding to the type and occupancy of the parking slot as

$$loss_{SCN} = w_{st} \cdot loss_{st} + w_{socc} \cdot loss_{socc} \quad (12)$$

where $w_{st}$, and $w_{socc}$ are the weights for the two losses and experimentally set.

The loss for the type of the parking slot that contains the center of the purple ROI in Fig. 7(d), $loss_{st}$ is calculated based on the categorical cross-entropy as

$$loss_{st} = \sum_{i=1}^{R/2} I_{slot}^i \left[ -\sum_{c=1}^{3} \{\lambda_{st,c} st_{true,c}^i \log(st_{pred,c}^i)\} \right] \quad (13)$$

where $st_{true,c}^i$ is the ground truth for the probability that the type of the parking slot containing the center of the $i$-th ROI is $c$. $st_{true}^i$ is represented in one-hot encoding and $c$ has a value of 1, 2, or 3. So $(st_{true,1}^i, st_{true,2}^i, st_{true,3}^i)$ for the perpendicular, parallel, or slanted type is set to $(1, 0, 0)$, $(0, 1, 0)$, or

(0, 0, 1), respectively. The number of ROIs for the SCN is $R/2$ when there are $R$ ROIs for the SDN because one region proposal contains one purple ROI and two magenta and yellow ROIs. $I_{slot}^i$ indicates whether the $i$-th ROI is included in a parking slot or not. Its value is set to 1 if included or 0 if not. $st_{pred,c}^i$ is the prediction of the network for $st_{true,c}^i$, $\lambda_{st,c}$ is the parameter that compensates for the imbalance of the numbers of different types of parking slots and is set based on the ratio of those numbers in the training dataset.

The loss for the occupancy of the parking slot that contains the center of the purple ROI in Fig. 7(d), $loss_{socc}$ is calculated as

$$loss_{socc} = \sum_{i=1}^{R/2} \left[ I_{occ}^i \left( socc_{pred}^i - socc_{true}^i \right)^2 + \lambda_{vac} I_{vac}^i \left( socc_{pred}^i - socc_{true}^i \right)^2 \right] \quad (14)$$

where $socc_{true}^i$ is the ground truth for the occupancy of the parking slot containing the center of the $i$-th ROI. This value is 1 if occupied or 0 if vacant. $socc_{pred}^i$ is the prediction of the network for $socc_{true}^i$. $I_{occ}^i$ indicates whether the center of the $i$-th ROI is included in an occupied parking slot and is set to 1 if included or 0 if not. $I_{vac}^i$ indicates whether the center of the $i$-th ROI is included in a vacant parking slot and is set to 1 if included or 0 if not. $\lambda_{vac}$ is the parameter that compensates for the imbalance of the numbers of the occupied and vacant parking slots. This value is set based on the ratio of those numbers in the training dataset.

## 4 Experiments

### 4.1 Dataset

The proposed method was quantitatively evaluated using two large-scale public parking slot detection datasets: Seoul National University dataset (Do & Choi, 2020) and Tongji Parking Slot Dataset 2.0 (Zhang et al., 2018). This paper will call them the SNU dataset and PS2.0 dataset, respectively. Table 1 shows the summary of the two datasets. The SNU dataset consists of half AVM images obtained by two fisheye cameras attached to both side-view mirrors. This dataset includes 22817 images (18299 for training and 4518 for test) taken in 571 parking situations, and the image resolution is 768×256 pixels that correspond to 14.4×4.8 meters. Its labels contain locations, orientations, types, and occupancies of the parking slots. On the other hand, the PS2.0 dataset consists of full AVM images obtained by four fisheye cameras of the AVM system. It includes 12165 images (9827 for training and 2338 for test) taken in 166 parking situations, and the image resolution is 600×600 pixels that correspond to 10.0×10.0 meters. Its labels contain only locations and orientations of the parking slots, so we manually designated their types and occupancies. The two datasets include three types of parking slots (perpendicular, parallel, and slanted) taken indoors and outdoors in daytime and nighttime under sunny and rainy weather conditions. Note that the SNU dataset is more challenging than the PS2.0 dataset mainly because of two reasons: One is that the SNU dataset contains a greater number of various parking situations, and the other is that the test and training images of the SNU dataset were taken from different parking situations while those of the PS2.0 dataset were taken from similar situations.

Table 1. Summary of the SNU and PS2.0 datasets.

|  |  | SNU dataset | PS2.0 dataset |
|---|---|---|---|
| Parking situations |  | 571 | 166 |
| Image resolution |  | 768 × 256 pixels (14.4 × 4.8 meters) | 600 × 600 pixels (10.0 × 10.0 meters) |
| Number of images | Training | 18299 | 9827 |
|  | Test | 4518 | 2338 |
|  | Total | 22817 | 12165 |
| Number of parking slots in training images | Perpendicular | 39743 | 5668 |
|  | Parallel | 5867 | 3492 |
|  | Slanted | 3276 | 316 |
|  | Total | 48886 | 9476 |
| Number of parking slots in test images | Perpendicular | 888 | 936 |
|  | Parallel | 11653 | 1151 |
|  | Slanted | 1004 | 81 |
|  | Total | 13545 | 2168 |

### 4.2 Experimental setting

The input images were resized to $576 \times 192$ pixels and $416 \times 416$ pixels for the SNU and PS2.0 datasets, respectively. The backbone network was initialized by the weights pre-trained on ImageNet, and the RPN, SDN, and SCN were initialized by Xavier uniform initializer. The proposed network was trained for 80 epochs with a batch size of 32. In the first 60 epochs, the first stage (RPN) and the second stage (SDN and SCN) were trained alternately for one epoch each, and in the rest 20 epochs, both stages were trained simultaneously. The proposed network was optimized by Adam optimizer whose learning rate, $\beta_1$, $\beta_2$, and $\epsilon$ were set to $10^{-4}$, 0.9, 0.999, and $10^{-8}$, respectively. Hyperparameters used to calculate losses are presented in Table 2. All the experiments were conducted using Tensorflow and Nvidia GeForce RTX 3090 GPU.

For proper evaluation and comparison, this paper utilizes the criteria suggested by Zhang et al. (2018), which is most widely used in previous parking slot detection papers. According to the criteria, a detected parking slot is considered as a true positive if the locations of its two junctions are within $M$ pixels from the ground truth and their orientations are within $N$ degrees from the ground truth. Otherwise, it is considered as a false positive. For $M$ and $N$, Zhang et al. (2018) used 12 pixels and 10 degrees (loose criteria), but this paper additionally uses 6 pixels and 5 degrees (tight criteria) for more detailed comparisons. Recall and precision are calculated as:

$$Recall = \frac{\#True\ Positive}{\#Ground\ Truth} \tag{15}$$

$$Precision = \frac{\#True\ Positive}{\#True\ Positive + \#False\ Positive} \tag{16}$$

Table 2. Hyperparameters used to calculate losses.

| | Parameter | SNU dataset | PS2.0 dataset |
|---|---|---|---|
| $loss_{first}$ | $w_{ep}, w_{exy}, w_{el}, w_{eo}, w_{so}$ | 400, 400, 1000, 1000, 400 | 500, 400, 1000, 1500, 500 |
| | $\lambda_e$ | 0.03 | 0.01 |
| | $L_{max}$ | 400 | 291 |
| $loss_{second}$ — $loss_{SDN}$ | $w_{jp}, w_{jxy}, w_{jo}$ | 1500, 2000, 6000 | 1000, 3000, 4000 |
| | $R$ | 12 | 8 |
| $loss_{second}$ — $loss_{SCN}$ | $w_{st}, w_{socc}$ | 0.5, 100 | 0.5, 100 |
| | $\lambda_{st,1}, \lambda_{st,2}, \lambda_{st,3}$ | 8.33, 1.23, 14.92 | 1.76, 2.86, 31.65 |
| | $\lambda_{vac}$ | 0.74 | 0.47 |

Table 3. Detection performances of the proposed method with different backbone networks.

| Backbone network | Recall | Precision |
|---|---|---|
| VGG16 | 91.28% | 91.78% |
| ResNet50 | 94.42% | 94.19% |
| **DenseNet121** | **95.75%** | **95.78%** |

### 4.3 Performance on the SNU dataset

This paper has considered several backbone networks and selected DenseNet121. Table 3 shows the detection performance of the proposed method with three different backbone networks: VGG16 (Simonyan & Zisserman, 2014), ResNet50 (He et al., 2016), and DenseNet121 (Huang et al., 2017). Since DenseNet121 outperforms the others, we utilized it to obtain the experimental results of the proposed method in the rest of this paper.

Table 4 presents the detection performances of the proposed method and two recently released methods. The two previous methods are the one-stage method by Suhr and Jung (2021) and the two-stage method by Do and Choi (2020). They were selected for this comparison because they achieved state-of-the-art performances on the PS2.0 and SNU dataset, respectively. In Table 4, the one-stage method (Suhr & Jung, 2021) shows a slightly higher performance than the two-stage method (Do & Choi, 2020). As mentioned in the introduction, it is mainly because the two-stage approach has not yet been adequately specialized for parking slot detection. It can be observed that the proposed method, a highly specialized two-stage parking slot detector, outperforms the others by roughly 3% to 5% with the loose criteria and by 11% to 13% with the tight criteria. This result signifies that the two-stage approach can outperform the one-stage approach in parking slot detection if it is well-specialized, which is the same as in the case of general object detection. In addition, when tightening the criteria, the performance of the proposed method drops only by about 12%, while those of the others dramatically drop by about 20%. This is primarily because the proposed method provides more accurate positions of the parking slots compared to the others. Table 5 gives the detailed positioning accuracies of the three methods. These errors were calculated from the correctly detected parking slots only. This table clearly shows that both the location and orientation errors of the proposed method are smaller than those of the others. In autonomous parking systems, positioning accuracy is significantly important because cars should be controlled based on the detected position of the parking slots.

Table 4. Comparison of parking slot detection performances on the SNU dataset.

| Method | Loose criteria (12 pixels, 10 degrees) | | Tight criteria (6 pixels, 5 degrees) | |
|---|---|---|---|---|
| | Recall | Precision | Recall | Precision |
| Proposed method | **95.75%** | **95.78%** | **83.14%** | **83.16%** |
| One-stage method by Suhr and Jung (2021) | 92.00% | 92.00% | 72.37% | 72.37% |
| Two-stage method by Do and Choi (2020) | 91.47% | 90.88% | 70.67% | 70.21% |

Table 5. Comparison of parking slot positioning accuracies on the SNU dataset.

| Method | Location error (pixel / cm) | | Orientation error (degree) | |
|---|---|---|---|---|
| | Mean | Std | Mean | Std |
| Proposed method | **2.12 / 5.30** | **1.39 / 3.48** | **1.12** | **1.06** |
| One-stage method by Suhr and Jung (2021) | 2.43 / 6.08 | 1.50 / 3.75 | 1.57 | 1.47 |
| Two-stage method by Do and Choi (2020) | 3.48 / 8.70 | 2.19 / 5.48 | 1.16 | 1.10 |

Table 6. Ablation experiment of the proposed method. (✓ indicates included)

| Method | | | Loose criteria (12 pixels, 10 degrees) | | Tight criteria (6 pixels, 5 degrees) | |
|---|---|---|---|---|---|---|
| | | | Recall | Precision | Recall | Precision |
| Case I | Region-specific ROIs | | 93.02% | 92.95% | 68.92% | 68.87% |
| | Multi-scale feature maps | | | | | |
| Case II | Region-specific ROIs | ✓ | 95.08% | 95.05% | 80.99% | 80.96% |
| | Multi-scale feature maps | | | | | |
| Case III | Region-specific ROIs | ✓ | **95.75%** | **95.78%** | **83.14%** | **83.16%** |
| | Multi-scale feature maps | ✓ | | | | |

Table 6 shows the result of the ablation experiment. Since this paper proposes the region-specific multi-scale feature extraction, this experiment was conducted focusing on the region-specific ROIs and multi-scale feature maps. In this table, from top to bottom, three cases present the detection results of using none of the region-specific ROIs and multi-scale feature maps; using only the region-specific ROIs without the multi-scale feature maps; and using both, respectively. In case I, the method designates the whole parking slot region as an ROI using a parallelogram as shown in Fig. 5(a). Compared to case I, with the tight criteria, case II reveals that the region-specific ROIs dramatically increase the detection performance by roughly 12%, and case III shows that the region-specific ROIs with the multi-scale feature maps enhance the detection performance by roughly 14%. The performances using the loose criteria have similar trends with smaller gaps. This ablation experiment clearly indicates that the proposed region-specific multi-scale feature extraction improves the parking slot detection performance.

Table 7 shows the type and occupancy classification performances of the three methods. Classification rates are also calculated from the correctly detected parking slots only. The type and occupancy classification rates of the proposed method all exceed 99%, and those of the other methods are quite similar. Table 8 presents the inference times of the three methods using Nvidia GeForce RTX 3090. The proposed method is faster than the two-stage method (Do & Choi, 2020) because its first and second stages share the same backbone network while Do and Choi's method uses two separate backbone networks. Compared to the one-stage method (Suhr & Jung, 2021), the proposed method is slower, as expected. This is consistent with the case of general object detection: the two-stage approach is slower but performs better than the one-stage approach.

Table 7. Comparison of type and occupancy classification performances on the SNU dataset.

| Method | Type classification rate | Occupancy classification rate |
|---|---|---|
| Proposed method | 99.92% | 99.07% |
| One-stage method by Suhr and Jung (2021) | 99.84% | 98.84% |
| Two-stage method by Do and Choi (2020) | 100% | 99.29% |

Table 8. Comparison of inference times using Nvidia GeForce RTX 3090.

| Method | Time (ms) | Frames per second | Framework |
|---|---|---|---|
| Proposed method | 22.11 | 45 | Tensorflow |
| One-stage method by Suhr and Jung (2021) | 14.10 | 71 | Tensorflow |
| Two-stage method by Do and Choi (2020) | 35.42 | 28 | Tensorflow |

Fig. 8 illustrates the parking slot detection results in various parking situations contained in the test images of the SNU dataset. In this figure, green, red, and blue lines indicate perpendicular, parallel, and slanted parking slots, respectively, and solid and dashed lines indicate vacant and occupied parking slots, respectively. It is apparent that the proposed method can successfully detect and classify parking slots under various illumination conditions (indoor, outdoor, nighttime, daytime, etc.), ground conditions (strong shadow, reflective floor, brick, grass, etc.) as well as parking slot styles (different colors, with marks, double separating line, etc.).

Fig. 9 presents failure cases of the proposed method in the test images of the SNU dataset. Figs. 9(a)-(c) show false positives. In Fig. 9(a), the lower junction of the detected parking slot does not satisfy the location criterion due to the reflective floor. In Fig. 9(b), the orientation of the detected parking slots does not satisfy the orientation criterion because of the image distortion. In Fig. 9(c), the rear part of the parking slot marking (rightmost blue line) is wrongly detected due to the shape similarity. Figs. 9(d) and (e) show incorrect classifications. In Fig. 9(d), the upper perpendicular parking slot is misclassified as a slanted parking slot. In Fig. 9(e), the parking slot occupied by a pole is misclassified as vacant because the pole occupies only a tiny area. Fig. 9(f) shows a false negative, where the upper junction of the parking slot is heavily occluded by the parked car.

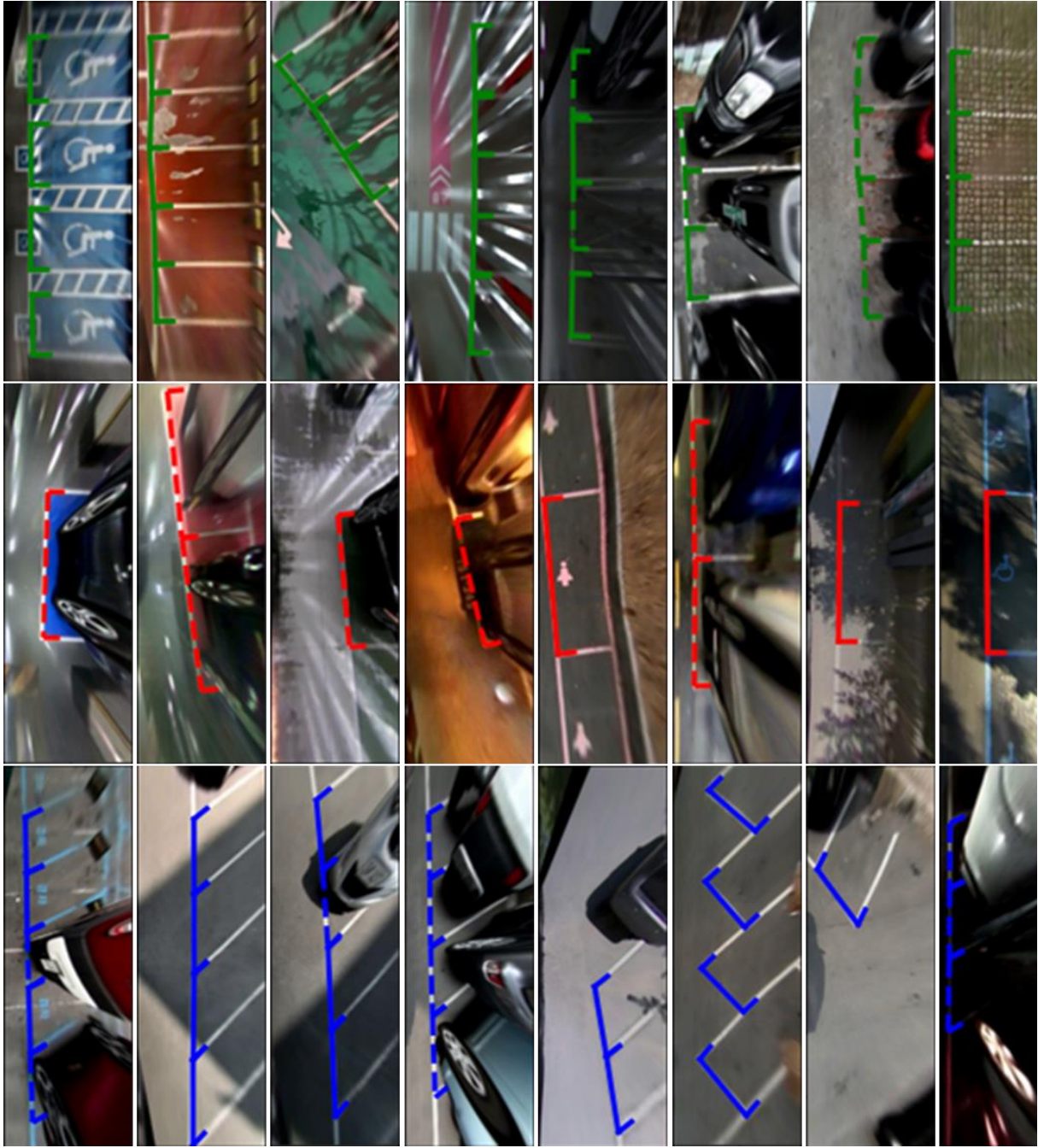

Fig. 8. Parking slot detection results of the proposed method in the test images of the SNU dataset. Green, red, and blue lines indicate perpendicular, parallel, and slanted parking slots, respectively; and solid and dashed lines indicate vacant and occupied parking slots, respectively.

### 4.4 Performance on the PS2.0 dataset

Table 9 shows the comparison of the parking slot detection performances on the PS2.0 dataset. For the PS2.0 dataset, three more methods have been added for the comparison because more papers shared their codes and detection results, unlike the newly opened SNU dataset. In Table 9, the proposed method shows a slightly higher parking slot detection performance than the others. Note that the performance gaps on the PS2.0 dataset are not as apparent as on the SNU dataset because almost all methods have already reached very high detection performances on this dataset. This is mainly due to the similarity between the training and test images of the PS2.0 dataset. This similarity makes it hard to be used to compare the performances of different methods. Table 10 compares the type and occupancy classification performances on the PS2.0 dataset. It also shows that the proposed method gives a slightly higher parking slot classification performance than the others. The previous methods with no ability for type or occupancy classification are masked as N/A.

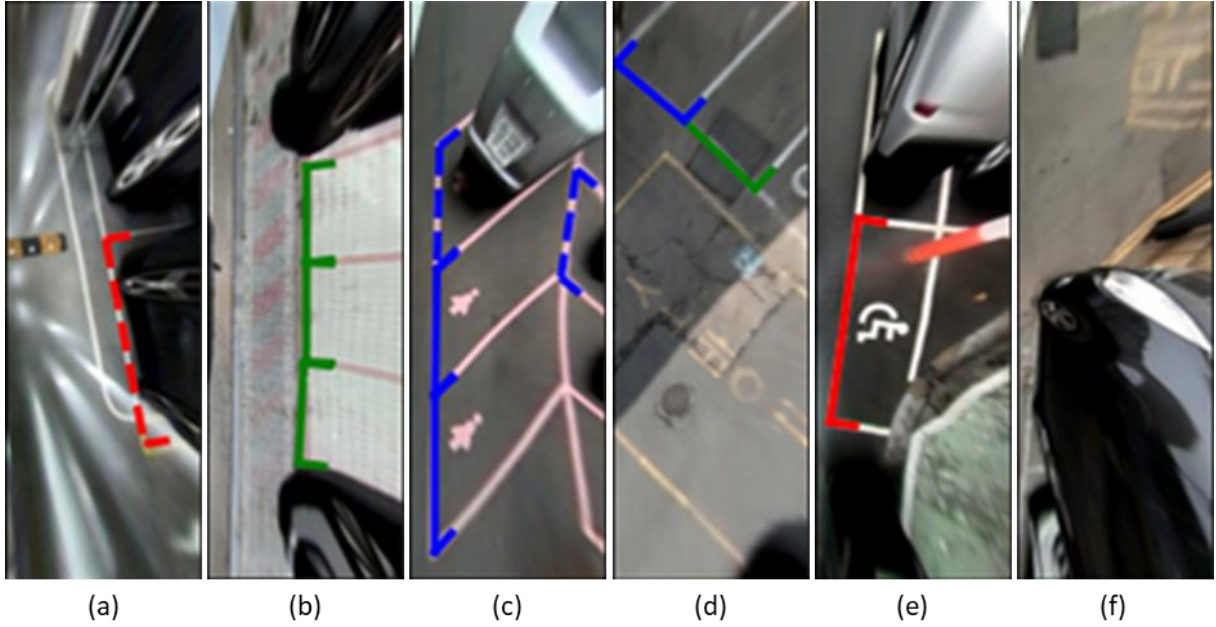

Fig. 9. Failure cases of the proposed method in the test images of the SNU dataset. (a), (b), and (c) show false positives, (d) and (e) show incorrect classifications, and (f) shows a false negative.

Table 9. Comparison of parking slot detection performances on the PS2.0 dataset.

| Method | Loose criteria (12 pixels, 10 degrees) | | Tight criteria (6 pixels, 5 degrees) | |
|---|---|---|---|---|
| | Recall | Precision | Recall | Precision |
| Proposed method | **99.77%** | **99.77%** | **99.54%** | **99.54%** |
| One-stage method (Suhr & Jung, 2021) | 99.77% | 99.77% | 99.45% | 99.45% |
| Two-stage method (Do & Choi, 2020) | 94.43% | 95.22% | 73.35% | 73.97% |
| VPS (Li, Cao, Yan, et al., 2020) | 99.31% | 99.40% | 99.22% | 99.17% |
| DMPR-PS (Huang et al., 2019) | 93.13% | 96.51% | 92.34% | 95.70% |
| DeepPS (Zhang et al., 2018) | 98.99% | 99.63% | 97.88% | 98.51% |

Table 10. Comparison of parking slot classification performances on the PS2.0 dataset.

| Method | Type classification rate | Occupancy classification rate |
|---|---|---|
| Proposed method | 100% | 99.40% |
| One-stage method (Suhr & Jung, 2021) | 100% | 99.31% |
| Two-stage method (Do & Choi, 2020) | 100% | 98.54% |
| VPS (Li, Cao, Yan, et al., 2020) | N/A | 98.33% |
| DMPR-PS (Huang et al., 2019) | N/A | N/A |
| DeepPS (Zhang et al., 2018) | N/A | N/A |

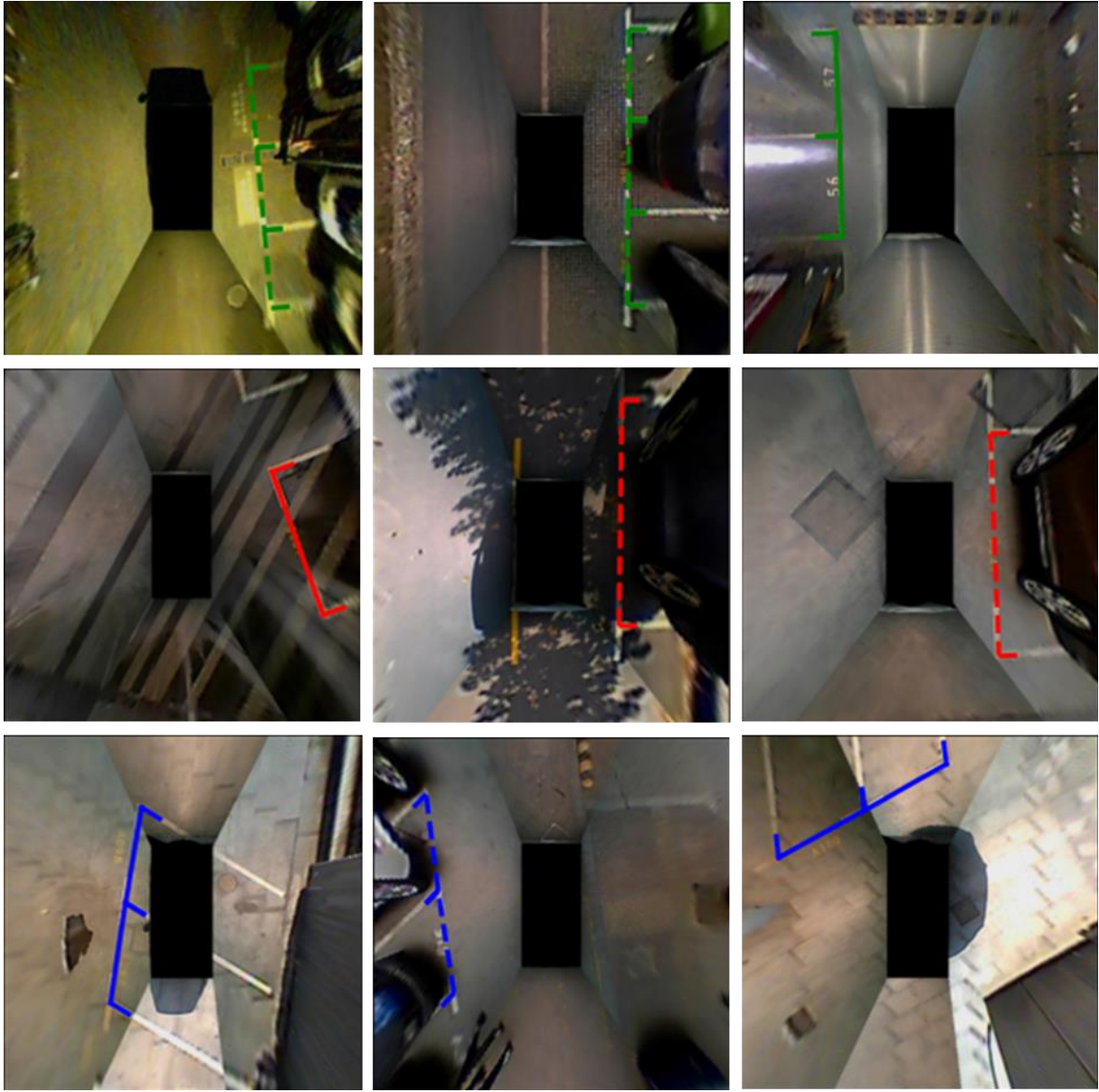

Fig. 10. Parking slot detection results of the proposed method in the test images of the PS2.0 dataset. Green, red, and blue lines indicate perpendicular, parallel, and slanted parking slots, respectively; and solid and dashed lines indicate vacant and occupied parking slots, respectively.

Fig. 10 illustrates the parking slot detection results in various parking situations contained in the test images of the PS2.0 dataset. It also shows that the proposed method can properly handle the various situations included in the PS2.0 dataset. Fig. 11 presents failure cases of the proposed method on the PS2.0 dataset. Fig. 11(a) shows a false positive where a space between two parking slots is mistakenly detected. Fig. 11(b) includes a false negative where the lower parking slot is undetected because one of its junctions is severely blurred. In Fig. 11(c), the occupied parking slot is misclassified as vacant.

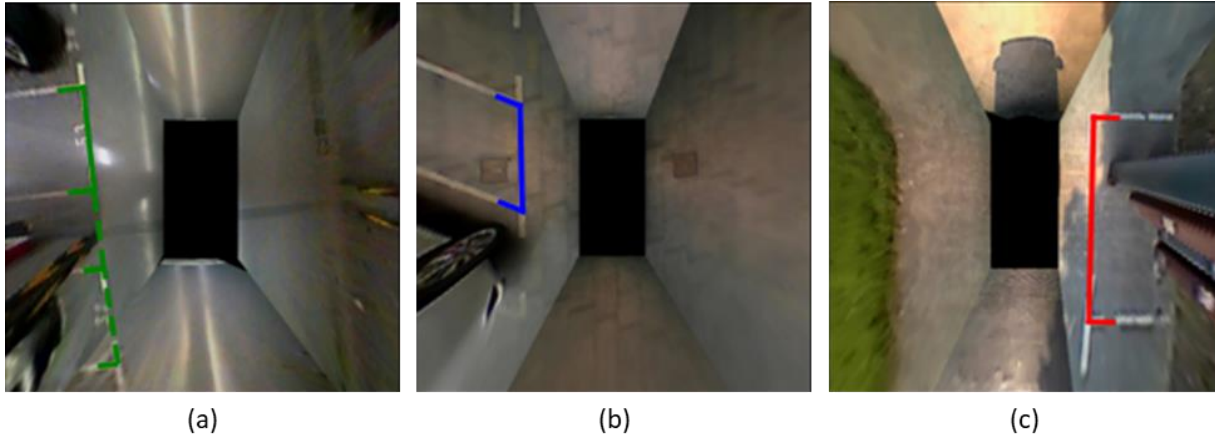

Fig. 11. Failure cases of the proposed method in the test images of the PS2.0 dataset. (a) shows a false positive, (b) shows a false negative, and (c) shows an incorrect classification.

## 5 Conclusion

This paper proposes a novel highly specialized two-stage parking slot detection method using the region-specific multi-scale feature extraction. The proposed method finds parking slot entrances as region proposals in the first stage and extracts region-specific features from multi-scale feature maps to precisely predict positions and properties of parking slots in the second stage. This method was quantitatively evaluated using two large-scale public parking slot detection datasets and outperformed previous methods in terms of both detection performance and positioning accuracy. This result revealed that the two-stage approach is superior to the one-stage approach if it is adequately specialized, the same as in the case of general object detection. In future, we are planning to optimize the network using filter pruning and weight quantization to implement it in real-time embedded systems and try to improve performance by integrating sequential detection results.